\documentclass[10pt,twocolumn,letterpaper]{article}

\usepackage{cvpr}              
\usepackage{amsmath,amssymb,amsfonts}
\usepackage{algorithmic}
\usepackage{graphicx}
\usepackage{textcomp}
\usepackage{booktabs}
\usepackage{multirow}
\usepackage{multicol}
\usepackage{caption}
\usepackage{adjustbox}
\usepackage{array}

\usepackage{colortbl}
\usepackage{times}
\usepackage{wrapfig}
\usepackage[accsupp]{axessibility}
\definecolor{cvprblue}{rgb}{0.21,0.49,0.74}
\usepackage[pagebackref,breaklinks,colorlinks,allcolors=cvprblue]{hyperref}

\def\paperID{15040} 
\def\confName{CVPR}
\def\confYear{2025}

\title{ODE: Open-Set Evaluation of Hallucinations in Multimodal Large Language Models}

\author{
Yahan Tu \quad Rui Hu \quad Jitao Sang\\
Beijing Jiaotong University\\
Beijing, China\\
{\tt\small \{yahan.tu, rui.hu, jtsang\}@bjtu.edu.cn}
}

\begin{document}
\maketitle
\begin{abstract}
Hallucination poses a persistent challenge for multimodal large language models (MLLMs). However, existing benchmarks for evaluating hallucinations are generally static, which may overlook the potential risk of data contamination. To address this issue, we propose ODE, an open-set, dynamic protocol designed to evaluate object hallucinations in MLLMs at both the existence and attribute levels. ODE employs a graph-based structure to represent real-world object concepts, their attributes, and the distributional associations between them. This structure facilitates the extraction of concept combinations based on diverse distributional criteria, generating varied samples for structured queries that evaluate hallucinations in both generative and discriminative tasks. Through the generation of new samples, dynamic concept combinations, and varied distribution frequencies, ODE mitigates the risk of data contamination and broadens the scope of evaluation. This protocol is applicable to both general and specialized scenarios, including those with limited data. Experimental results demonstrate the effectiveness of our protocol, revealing that MLLMs exhibit higher hallucination rates when evaluated with ODE-generated samples, which indicates potential data contamination. Furthermore, these generated samples aid in analyzing hallucination patterns and fine-tuning models, offering an effective approach to mitigating hallucinations in MLLMs. Our code are available at \href{https://github.com/Iridescent-y/ODE}{https://github.com/Iridescent-y/ODE}.
\end{abstract}   
\section{Introduction}
\label{sec:intro}

Multimodal Large Language Models (MLLMs) \cite{Gpt4, Liu_2024_CVPR, InstructBLIP, zhu2023minigpt, ye2023mplug, wang2023cogvlm} have rapidly advanced in recent times, enabling detailed image descriptions (i.e., image captioning) and responses to image-related queries (i.e., visual question answering). However, these models face the persistent challenge of ``hallucination'' \cite{liu2023aligning, huang2023survey, wang2024valid}, where generated responses appear plausible but lack fidelity to the actual image content. This issue can lead to harmful consequences, limiting the utility of MLLMs.

\begin{figure}[!tb]
  \centering
  \centerline{\includegraphics[width=\linewidth]{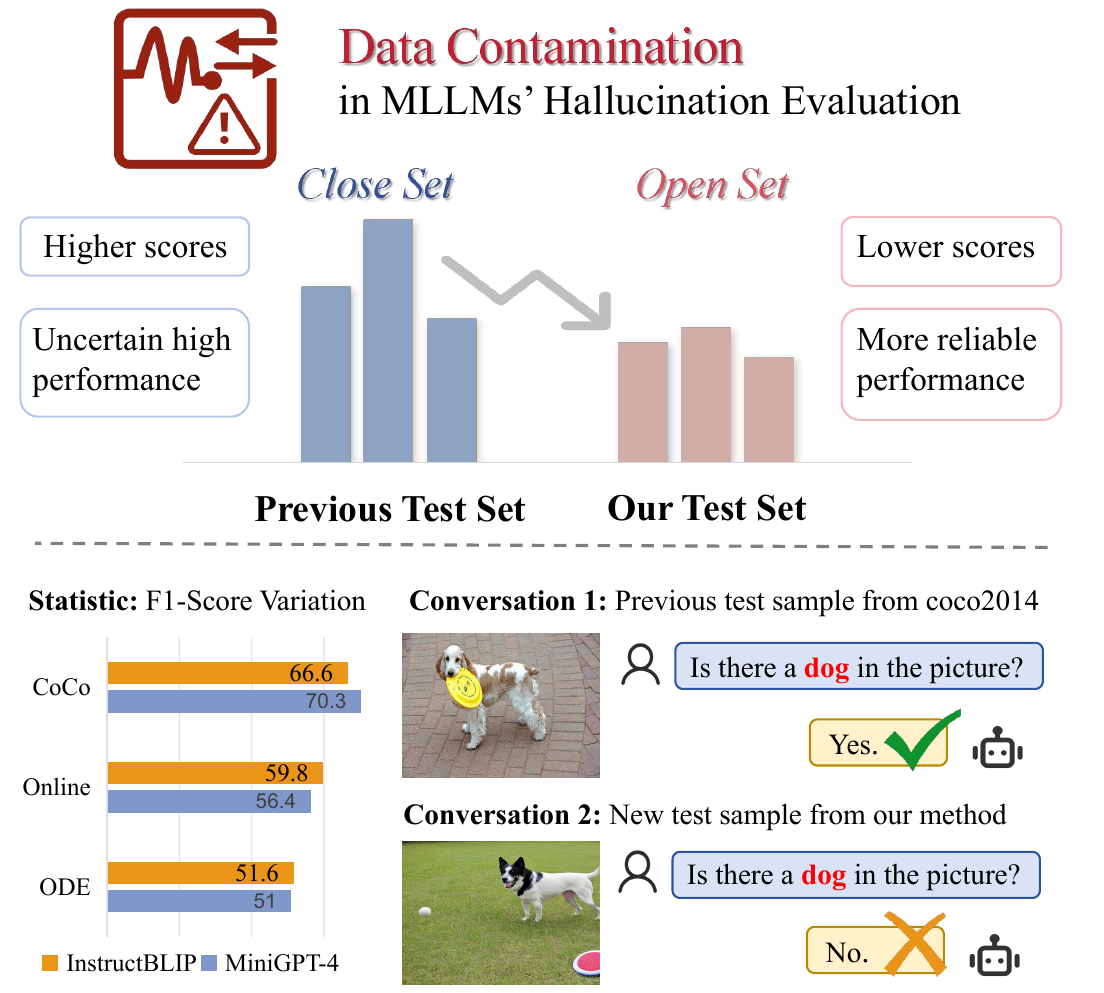}}
\caption{Comparison of closed-set and open-set evaluations for MLLMs, showing that open-set testing reduces data contamination and provides a more reliable assessment of hallucination rates.}
\label{fig:fig1}
\end{figure}

Therefore, the evaluation of hallucinations in MLLMs is crucial to improve model reliability and practical application. Prior studies have proposed various benchmarks to evaluate hallucinations in MLLMs, focusing on different types (e.g., existence hallucinations\cite{li2023evaluating, wang2023llm} and relational hallucinations \cite{zheng2024reefknot}) or levels of difficulty \cite{pmlr-v235-wu24l, HallusionBench, HaloQuest}. However, these benchmarks are predominantly static, using fixed test data with limited distributions, which increases the risk of data contamination. Contamination occurs when test data overlaps with training data, leading to inflated performance metrics. For example, we find that, under the same semantic distribution, models perform better on the COCO2014 image subset than on the latest Internet images (as shown in Fig.~\ref{fig:fig1}), with the latest Internet data being less likely to be contaminated by training overlap, thus providing a more reliable evaluation. This raises concerns about whether correct responses on COCO reflect genuine understanding or result from data contamination.

Recent studies\cite{zhou2023don, li2023open} highlight the issue of data contamination in Large Language Models (LLMs). Some LLM studies introduce dynamic evaluation methods to address this risk. For instance, DyVal \cite{zhu2023dyval} dynamically synthesizes test samples using a directed acyclic graph, though its application is limited to specific algorithms. Similarly, MSTemp\cite{liu2023meta} generates semantically equivalent evaluation samples based on the SST-2 dataset, yet its scope remains confined to the distribution of that dataset. To the best of our knowledge, there is currently no evaluation method existing to specifically mitigate data contamination in MLLMs.

We argue that an effective evaluation benchmark should be open-set, meaning that evaluation data are novel to the model at both sample and distribution levels. Inspired by contamination studies in the LLM domain and guided by insights into the unique challenges of multimodal models, we outline three distinctive features that characterize our approach: (1) Out-of-distribution (OOD) evaluation at a broad distributional level rather than modifying existing datasets; (2) dynamic sample generation across different modalities; and (3) a multi-tiered dynamic structure extending from concept-level to attribute-level and distribution-level granularity.

Building on these characteristics, we introduce the Open-Set Dynamic Evaluation (ODE) protocol. ODE automatically generates datasets to evaluate object hallucinations in MLLMs, covering both existence-level and attribute-level hallucinations. Initially, we model real-world object concepts, object attributes, and concept-attribute combinations as a graph. Then, from this graph, we extract concept nodes and their associated attributes, designing diverse semantic scenes and queries for each test sample and synthesizing high-quality images. To guide the selection of concept node pairs, we design four frequency-based criteria in the following order: Standard, Long-tail, Random, and Fictional. Each criterion reflects a unique distribution pattern of object combination frequencies (see Section 2.2 for detailed explanation). 

The ODE protocol enables iterative dynamic updates to the dataset, generating new content based on selected concepts and their distribution combinations. The automated protocol enhances controllability while reducing human intervention. The phenomenon in Fig.~\ref{fig:fig1}, where the performance of our synthetic images is comparable to internet-sourced images, along with subsequent experiments, validates the effectiveness of synthetic images. We conducted extensive evaluations on multiple MLLMs under varied criteria, finding that hallucination rates were more pronounced than on existing static benchmarks, with performance differences observed across distributions, tasks, and models. Further analysis revealed varying hallucination tendencies linked to specific concepts. Finally, we conduct fine-tuning experiments using ODE-generated samples, demonstrating that selectively refining models on ODE-identified error samples or directly fine-tuning on ODE-generated data effectively reduces hallucinations. Our findings primarily illustrate ODE’s utility in general scenarios; additionally, ODE can also be applied to domains with data scarcity or imbalanced distributions.

In summary, this paper makes the following contributions:
\begin{itemize}
    \item We introduce an open-set dynamic evaluation protocol that generates novel samples through dynamic target concept combinations, mitigating potential data contamination in MLLM hallucination evaluation.
    \item We perform extensive hallucination evaluations across multiple models, showing the effectiveness of our protocol.
    \item The generated evaluation data aid model debugging, and targeted fine-tuning on ODE-identified errors or general ODE-generated samples effectively enhances model performance.
\end{itemize}
\section{Related Work}
\label{sec:relatedwork}

\subsection{Hallucination Evaluation in MLLMs}

\begin{figure*}[!ht]
  \centering
  \includegraphics[width=\linewidth]{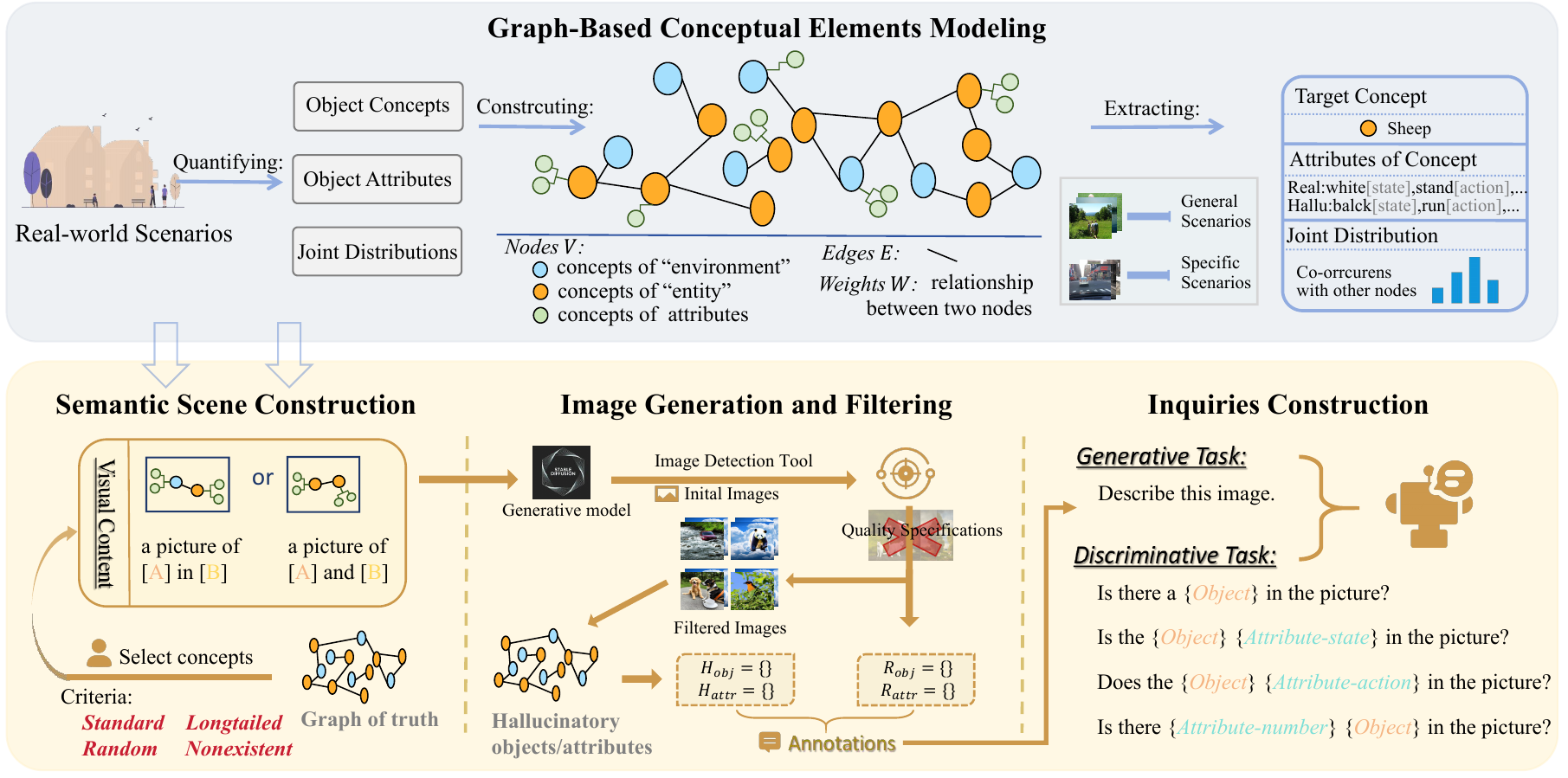}
   \caption{Pipeline of the Open-Set Dynamic Evaluation Protocol. The workflow involves constructing a graph and generating test samples based on the graph, with four distinct steps.}
  \label{fig:method}
\end{figure*}

To evaluate the degree of hallucination in MLLMs, various hallucination benchmarks have been proposed. The earliest proposed CHAIR \cite{CHAIR} measures the accuracy of object references in captions by calculating the precision of hallucinatory objects. POPE \cite{li2023evaluating} improves CHAIR by evaluating the presence of hallucinations based on object detection, suitable for discriminative tasks. AMBER \cite{wang2023llm} evaluates object hallucinations in both discriminative and generative tasks from three dimensions: existence, attributes, and relationships. Expanding beyond object hallucinations, examples include HallusionBench \cite{HallusionBench}, which focuses on visual common sense and reasoning; Hal-Eval \cite{haleval}, which examines event hallucinations; and CorrelationQA \cite{CorrelationQA}, which studies the impact of false visual inputs and other cues like counting and OCR. Some studies further evaluate complex hallucinations. VHTest \cite{VHTest} compares embedding similarity to find potential VH instances, while HaloQuest \cite{HaloQuest} uses real and synthesized generated images to focus on more complex scenes. Despite the proliferation of benchmarks, from simple to complex tasks, and small to large scales, testing within closed sets remains unsolved. We dynamically update the dataset by combining target concepts to achieve an open set of samples and distributions.

\subsection{Data Contamination}
Data contamination has attracted considerable attention in LLMs. The GPT-4\cite{gptperform} and LLama \cite{touvron2023llama} reports highlight this phenomenon. Zhou et al. \cite{zhou2023don} discussed the risks and impacts of benchmark data contamination in evaluating LLMs; similarly, Ni et al. \cite{ni2024training} explored related concerns. Several researchers have developed methods for detecting data contamination. Fan\cite{fan2023nphardeval}, Lei \cite{s3eval}, Zhu \cite{zhu2023dyval}, and others introduced dynamic evaluation strategies through different algorithms to reduce data pollution. The dataset generated through the ODE protocol is dynamically distributed and can be continuously updated, which avoids the problem of data pollution that may exist in existing static datasets for illusion evaluation.

\subsection{Dynamic Evaluation}

Dynamic Evaluation refers to dynamically generating evaluation data for testing. The key to early works such as DynaBoard \cite{Dynaboard}, DynaTask \cite{Dynatask}, and Dynabench \cite{Dynabench} was to utilize the intelligence of the crowd to challenge the design of evaluation sets, with the main effort focused on crowdsourcing systems and interfaces rather than algorithms, which required significant costs. Recent methods automate dynamic evaluation by rewriting and expanding samples based on task scenarios or datasets. Zhu et al. \cite{zhu2023dyval} used mathematical reasoning tasks to reduce data pollution in LLM by dynamically generating test samples from directed acyclic graphs. MsTemp \cite{MStemp}, DyVal2 \cite{zhu2024dyval}, and Benchmark self evolving\cite{wang2024benchmark} dynamically reconstruct samples from existing datasets. DeNEVIL \cite{DeNEVIL} iteratively updates and improves prompts based on values, and dynamically mines samples. Our approach reduces reliance on fixed datasets by dynamically combining meta-concepts and attributes, adapting visual content and textual prompts.

\section{Methodology}

This section outlines the ODE protocol for dynamically generating image content and prompt text for test data. As shown in Fig.~\ref{fig:method}, the workflow consists of four steps: modeling real-world scenarios using a graph structure, conceptual design of semantic scene, image generation and filtering, and template design for inquiries.

\subsection{Graph-Based Conceptual Modeling}

Aiming to cover a broader range of target object concepts in our evaluation of object existence hallucination, ODE employs a weighted graph \( G \) to model real-world scenes, facilitating the generation of more diverse scenarios in subsequent stages. Specifically, the graph \( G = (V, A, E, W) \) provides an abstract representation of real-world objects, their interrelations, and attributes. Common object categories are extracted from existing datasets to form the nodes \( V \), termed meta-concepts. In subsequent processes, a subset of these nodes will be selected as target concepts to test hallucination cases in the model. The attribute nodes \( A \) of the main nodes \( V \) represent the properties of each object concept, including state and action attributes. Additionally, quantity attributes are considered during testing. The edge weights \( W \) denote the strength of the relationships between nodes, based on the co-occurrence quantity of object concepts in real-world scenes. We consider that co-occurrence frequency can reflect the semantic association and distributional characteristics between the conceptual entities \( V_i \) and \( V_j \). For example, the high co-occurrence frequency between ''table'' and ''chair'' reflects their common pairing in indoor scenes. An edge \( E \) is established between nodes if a connection exists (i.e., the edge weight \( W \) is non-zero). To facilitate a more detailed analysis of the mutual influence of hallucinations among these concepts, we categorize the concepts into environment-level \( V_{\text{env}} \) and entity-level \( V_{\text{ent}} \), such as ''grass'' and ''frisbee''. We particularly focus on two coexistence patterns: entity-environment and entity-entity coexistence patterns, ensuring the content generated in subsequent stages targeted.

ODE performs concept extraction from real-world scenarios and then focuses on hallucinatory scenarios by generating a graph that captures associations related to hallucinations, facilitating the creation of a comprehensive range of test samples for general hallucination evaluation. This modeling approach is applicable to both general and specific domains. We further emphasize the customization of hallucination detection tailored to specific fields.

\subsection{Construction of Semantic Scenes}
\label{sec:conceptual-composition}
After obtaining a scene graph with object concepts, we select two concept nodes at each step to form a pair, which is used as the content for the test image. This image is then generated using a text-to-image model. 

\subsubsection{Selection Criteria}

Through extensive learning from multimodal data, the model gradually acquires semantic representations of concepts, however, it may show varying comprehension of concept combinations, especially with diverse distributions, potentially causing hallucinations. Based on this observation, we design four distributional levels of concept combination standards—\textbf{Standard}, \textbf{Long-tail}, \textbf{Random}, and \textbf{Fictional}—to evaluate the model's grasp of different association patterns.

In implementation, we first select two object concepts \( V_i \) and \( V_j \) from the graph structure and assign each object specific attributes, including state \( A_\text{state} \), action \( A_\text{action} \), and quantity \( A_\text{number} = 1 \) attributes. For Standard and Long-tail pairs, attributes are chosen from frequently observed categories, reflecting realistic scenarios, while attributes for Random and Fictional pairs are randomly chosen to increase diversity. The four selection criteria are defined as follows:

\begin{itemize}
    \item \textbf{Standard}: Pairs with the highest co-occurrence frequency:
    \begin{equation}
        (V_i, V_j) \in \text{argmax}_{i,j} \, c_{i,j} 
        \label{eq:1argmax}
    \end{equation}
    where \( c_{i,j} \) represents the co-occurrence count of objects \( V_i \) and \( V_j \) in the existing dataset. These combinations, with attributes from standard categories, test the model's understanding of high-frequency associations.

\begin{figure}[htb]
  \centering
  \centerline{\includegraphics[width=\linewidth]{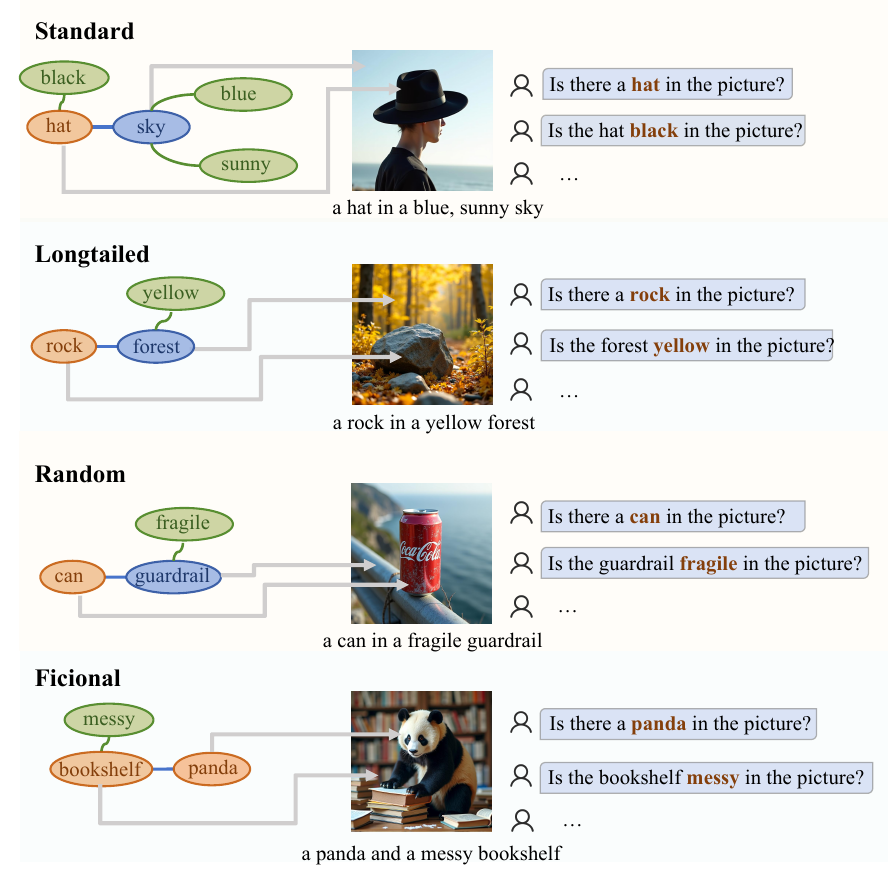}}
\caption{Examples of four distribution samples constructed by our method.}
\label{fig:example}
\end{figure}

    \item \textbf{Long-tail}: Pairs with moderate co-occurrence, simulating long-tail distributions:
    \begin{equation}       
    (V_i, V_j) \in \left\{ (V_k, V_l) \, \big| \, \epsilon < c_{k,l} < \delta \right\}
    \label{eq:2longtail}
    \end{equation}
    where \( \epsilon \) and \( \delta \) set co-occurrence bounds. This standard evaluates the performance of the model in rare, associative pairs. Attributes are also chosen from commonly observed attributes to align with realistic contexts.
    \item \textbf{Random}: Randomly selected pairs without co-occurrence considerations:
    \begin{equation}
    (V_i, V_j) \sim \text{Uniform}(V \times V)
    \label{eq:3random}
    \end{equation}
    Attributes are also randomly chosen, evaluating the model’s robustness to different semantic levels.

    \item \textbf{Fictional}: Pairs without any co-occurrence record:
    \begin{equation}
    (V_i, V_j) \notin \{ (V_k, V_l) \, | \, c_{k,l} > 0 \}
    \label{eq:4fictional}
    \end{equation}
    Fictional pairs assess the model’s reasoning with novel distributions, assigning attributes at random.
\end{itemize}

Using these criteria, concept pairs \( (V_i, V_j) \) and attributes are dynamically selected to generate test images across distributions, ensuring varied semantic scenarios in the dataset and supporting a thorough evaluation of the model’s semantic comprehension.

\subsubsection{Scene Content}

The semantic scene for each sample is constructed by extracting concept pairs and dynamically generating content, ensuring distinct test instances through inherent randomness. Our evaluation set includes two main object concepts and their attributes (or sometimes without specific attributes) in two combination types. The first type pairs two entity categories, requiring the model to capture direct interactions, such as ''a black running dog and a yellow frisbee.'' The second type pairs an entity with an environmental category, prompting the model to understand contextual relationships, such as ''a yellow deer and a lush forest'' This approach broadens the range of concepts, facilitating detailed classification analysis of hallucination tendencies. Fig.~\ref{fig:example} shows examples from the evaluation set.

\subsection{Generation and Filtering of Images}
To avoid model exposure to test data, we use text-to-image generation models (e.g., FLUX.1-dev and Stable Diffusion 1.5, as used in our experiments) to generate ODE test images from textual prompts like ``a picture of {attribute of A} A and {attribute of B} B,'' where A and B represent specific visual concepts. Positive prompts include “clear and obvious entity,” “photography,” and “concise,” while negative prompts use “bad anatomy,” “incomplete body,” and “mutated” to enhance quality.

Not all generated images meet quality standards due to generative model limitations. For each test case, we set different random seeds to produce multiple image variations with the same semantic scene, selecting higher-quality samples afterward. We assess image quality using an open vocabulary object detection model, discarding images lacking expected entities. For example, if an image described as ``a picture of a dog and a frisbee'' does not meet the required entities' confidence scores (below 0.65), it is filtered out. Our concept list includes all objects present in the images. By filtering low-confidence images, we retain high-quality samples, using the remaining concepts as ''ground truth'' data and annotating potential hallucinated objects based on object co-occurrence frequency.

\subsection{Structuring of Inquiries}

We develop an evaluation Inquiry template specifically designed to evaluate object-level and attribute-level hallucinations (state, action, and quantity) through automated generation.

For generative tasks, we use the inquiry ``Please describe this image.'' to instruct the MLLM to identify the image’s concepts. For discriminative tasks, object existence hallucinations are evaluated with inquiries like ``Is there a \{object\} in the image?'' Attribute hallucinations are evaluated with inquiries like ``Is the \{object\} \{state attribute\} in the image?'', ''Does the \{object\} \{action attribute\} in the picture?'', and ''Is/Are there \{quantity\} \{object(s)\}  in the picture?'', expecting a ``yes'' or ``no'' response. To examine hallucinated objects, we include counterfactual prompts in the discriminative task, inquiring about nonexistent objects or attributes, such as ``Is there a \{hallucinated object\} in the image?''
\definecolor{lightblue}{RGB}{200, 215, 235} 
\definecolor{deepblue}{RGB}{70, 110, 140} 
\definecolor{lightyellow}{RGB}{250, 235, 199} 
\definecolor{deepyellow}{RGB}{200, 150, 100} 
\definecolor{lightgray}{gray}{0.9}
\begin{table*}[ht]
    \centering
    \setlength{\tabcolsep}{7pt} 
    \caption{Evaluation results of different models on both generative and discriminative tasks across various scenarios. Highlighted cells indicate optimal performance. Yellow highlights denote better performance with lower values, while blue highlights indicate better performance with higher values.}
    \renewcommand{\arraystretch}{1.2} 
    \begin{adjustbox}{max width=\textwidth,center}
    \fontsize{9pt}{10pt}\selectfont
    \begin{tabular}{p{0.9cm}p{0.4cm}|l|p{0.9cm}p{0.7cm}p{0.7cm}p{0.7cm}|p{0.7cm}p{0.7cm}p{0.7cm}p{0.7cm}|p{0.7cm}p{0.7cm}p{0.7cm}p{0.7cm}} 
        \toprule
        \multicolumn{3}{c|}{} & \multicolumn{4}{c|}{\textbf{Generative Task}} & \multicolumn{4}{c|}{\textbf{Discriminative-Existence Task}} & \multicolumn{4}{c}{\textbf{Discriminative-Attribute Task}} \\
         \cmidrule(r){4-7} \cmidrule(l){8-11} \cmidrule(l){12-15}
        \multicolumn{2}{>{\centering\arraybackslash}m{2.75cm}|}{\textbf{Criterion}}  & \textbf{Model} & \textbf{CHAIR} & \textbf{Cover} & \textbf{Hal} & \textbf{Cog} & \textbf{Acc} & \textbf{P} & \textbf{R} & \textbf{F1} & \textbf{Acc} & \textbf{P} & \textbf{R} & \textbf{F1} \\
        \midrule
        \multicolumn{2}{>{\centering\arraybackslash}m{2.75cm}|}{\multirow{5}{*}{\centering AMBER}} & CogVLM & \cellcolor{lightyellow!30}\textbf{\textcolor{deepyellow}{4.0}} & 52.6 & \cellcolor{lightyellow!30}\textbf{\textcolor{deepyellow}{16.1}} & \cellcolor{lightyellow!30}\textbf{\textcolor{deepyellow}{1.2}} & 20.9 & \cellcolor{lightblue!30}\textbf{\textcolor{deepblue}{100.0}} & 20.9 & 34.5 & 45.4 & \cellcolor{lightblue!30}\textbf{\textcolor{deepblue}{92.1}} & 17.7 & 29.7\\
        & & LLaVA-1.5 & 7.7 & 50.4 & 34.4 & 3.7 & 71.0 & \cellcolor{lightblue!30}\textbf{\textcolor{deepblue}{100.0}} & 71.0 & 83.0 & \cellcolor{lightblue!30}\textbf{\textcolor{deepblue}{72.5}} & 89.8 & \cellcolor{lightblue!30}\textbf{\textcolor{deepblue}{50.7}} & \cellcolor{lightblue!30}\textbf{\textcolor{deepblue}{64.8}}\\
        & & mPLUG & 23.9 & 47.7 & 79.3 & 12.8 & 15.0 & \cellcolor{lightblue!30}\textbf{\textcolor{deepblue}{100.0}} & 15.0 & 26.1 & 49.3 & 64.5 & 17.4 & 27.4\\
        & & MiniGPT-4 & 17.9 & \cellcolor{lightblue!30}\textbf{\textcolor{deepblue}{61.2}} & 70.2 & 13.4 & \cellcolor{lightblue!30}\textbf{\textcolor{deepblue}{96.9}} & \cellcolor{lightblue!30}\textbf{\textcolor{deepblue}{100.0}} & \cellcolor{lightblue!30}\textbf{\textcolor{deepblue}{96.9}} & \cellcolor{lightblue!30}\textbf{\textcolor{deepblue}{98.4}} & 62.9 & 68.2 & 48.5 & 56.6\\
        & & InstructBLIP & 12.5 & 57.5 & 63.2 & 8.5 & 67.4 & \cellcolor{lightblue!30}\textbf{\textcolor{deepblue}{100.0}} & 67.4 & 80.5 & 70.6 & 88.8 & 47.2 & 61.6\\
        \midrule
        \multirow{10}{*}{\textbf{Ours}} & \multirow{5}{*}{Standard} & CogVLM & 51.9 & 76.5 & 89.1 & 12.2 & 50.7 & \cellcolor{lightblue!30}\textbf{\textcolor{deepblue}{100.0}} & 26.2 & 41.5 & 55.2 & \cellcolor{lightblue!30}\textbf{\textcolor{deepblue}{46.8}} & 55.6 & \cellcolor{lightblue!30}\textbf{\textcolor{deepblue}{50.8}}\\
        & & LLaVA-1.5 & \cellcolor{lightyellow!30}\textbf{\textcolor{deepyellow}{38.9}} & \cellcolor{lightblue!30}\textbf{\textcolor{deepblue}{77.7}} & \cellcolor{lightyellow!30}\textbf{\textcolor{deepyellow}{82.7}} & \cellcolor{lightyellow!30}\textbf{\textcolor{deepyellow}{8.6}} & \cellcolor{lightblue!30}\textbf{\textcolor{deepblue}{69.5}} & 97.8 & \cellcolor{lightblue!30}\textbf{\textcolor{deepblue}{55.4}} & \cellcolor{lightblue!30}\textbf{\textcolor{deepblue}{70.7}} & 62.9 & 32.6 & \cellcolor{lightblue!30}\textbf{\textcolor{deepblue}{71.9}} & 44.8\\
        & & mPLUG & 50.8 & 77.2 & 96.0 & 11.5 & 41.7 & 94.7 & 13.4 & 23.5 & \cellcolor{lightblue!30}\textbf{\textcolor{deepblue}{72.5}} & 41.7 & 28.6 & 33.9\\
        & & MiniGPT-4 & 49.4 & 76.0 & 93.6 & 14.2 & 64.5 & 97.5 & 48.0 & 64.3 & 69.5 & 39.7 & 12.5 & 19.0\\
        & & InstructBLIP & 59.9 & 75.7 & 88.1 & 11.0 & 66.7 & 96.8 & 51.7 & 67.4 & 60.8 & 28.5 & 51.2 & 36.6\\
        \cmidrule(r){2-15}
        & \multirow{5}{*}{Random} & CogVLM & 58.1 & 57.7 & 87.6 & 6.0 & 40.0 & \cellcolor{lightblue!30}\textbf{\textcolor{deepblue}{89.7}} & 18.0 & 30.0 & 57.1 & 82.5 & 38.2 & 52.2 \\
        & & LLaVA-1.5 & \cellcolor{lightyellow!30}\textbf{\textcolor{deepyellow}{45.2}} & 57.7 & 84.2 & \cellcolor{lightyellow!30}\textbf{\textcolor{deepyellow}{4.7}} & \cellcolor{lightblue!30}\textbf{\textcolor{deepblue}{74.7}} & \cellcolor{lightblue!30}\textbf{\textcolor{deepblue}{89.7}} & \cellcolor{lightblue!30}\textbf{\textcolor{deepblue}{69.6}} & \cellcolor{lightblue!30}\textbf{\textcolor{deepblue}{78.3}} & \cellcolor{lightblue!30}\textbf{\textcolor{deepblue}{75.7}} & 90.1 & \cellcolor{lightblue!30}\textbf{\textcolor{deepblue}{65.9}} & \cellcolor{lightblue!30}\textbf{\textcolor{deepblue}{76.1}}\\
        & & mPLUG & 57.9 & 56.4 & 92.1 & 6.3 & 40.0 & 84.0 & 10.8 & 19.1 & 48.4 & 76.5 & 20.2 & 31.9\\
        & & MiniGPT-4 & 50.3 & \cellcolor{lightblue!30}\textbf{\textcolor{deepblue}{57.9}} & \cellcolor{lightyellow!30}\textbf{\textcolor{deepyellow}{82.2}} & 5.8 & 66.4 & 86.9 & 58.2 & 69.7 & 45.0 & 74.5 & 10.8 & 18.8 \\
        & & InstructBLIP & 55.9 & 58.4 & 83.2 & 5.6 & 64.6 & 87.6 & 54.6 & 67.2 & 72.3 & \cellcolor{lightblue!30}\textbf{\textcolor{deepblue}{91.5}} & 58.6 & 71.4\\
        \bottomrule
    \end{tabular}%
    \end{adjustbox}
    \label{table:evaluation}
\end{table*}

\section{Experiments}

\subsection{Setup}

\textbf{Data Preparation.} 
We extracted real-world target concepts through the AMBER benchmark\cite{wang2023llm}, a multidimensional benchmark for evaluating MLLM hallucinations that includes 337 objects across 14 domains reflecting standard concepts. After concept modeling, we selected 40 concept combinations at four different distributional difficulty levels to generate images, resulting in 808 test images after sampling. Each image was paired with factual and hallucination questions, creating 8,786 test data pairs for both discriminative and generative tasks.

\textbf{MLLMs Evaluated.} We selected several state-of-the-art MLLMs for evaluation, including MiniGPT-4 \cite{zhu2023minigpt}, InstructBLIP \cite{InstructBLIP}, LLaVA-1.5 \cite{Liu_2024_CVPR}, CogVLM \cite{wang2023cogvlm}, mPLUG\_Owl \cite{ye2023mplug}, Cambrian \cite{tong2024cambrian}, InternVL-2.5 \cite{chen2024expanding}, and Qwen2-VL \cite{wang2024qwen2}. To ensure fair evaluation, we used each model's official hyperparameters to avoid length-based bias in response generation.

\textbf{Evaluation Metrics.} We adopted AMBER-derived evaluation metrics. For generative tasks, CHAIR measured hallucination frequency, Cover assessed content coverage, Hal represented the proportion of hallucinated responses, and Cog evaluated the similarity between generated hallucinations and human cognitive hallucinations. For discriminative tasks, we used standard classification metrics: Accuracy, Precision, Recall, and F1-Score. To evaluate hallucinations, ODE calculates Precision and Recall specifically for hallucination-related questions (ground truth ''no'') and uses Accuracy across all questions to prevent models from achieving high scores by simply rejecting answers.

\begin{figure}[htb]
    \centering
    \includegraphics[width=0.85\linewidth, height=4.8cm]{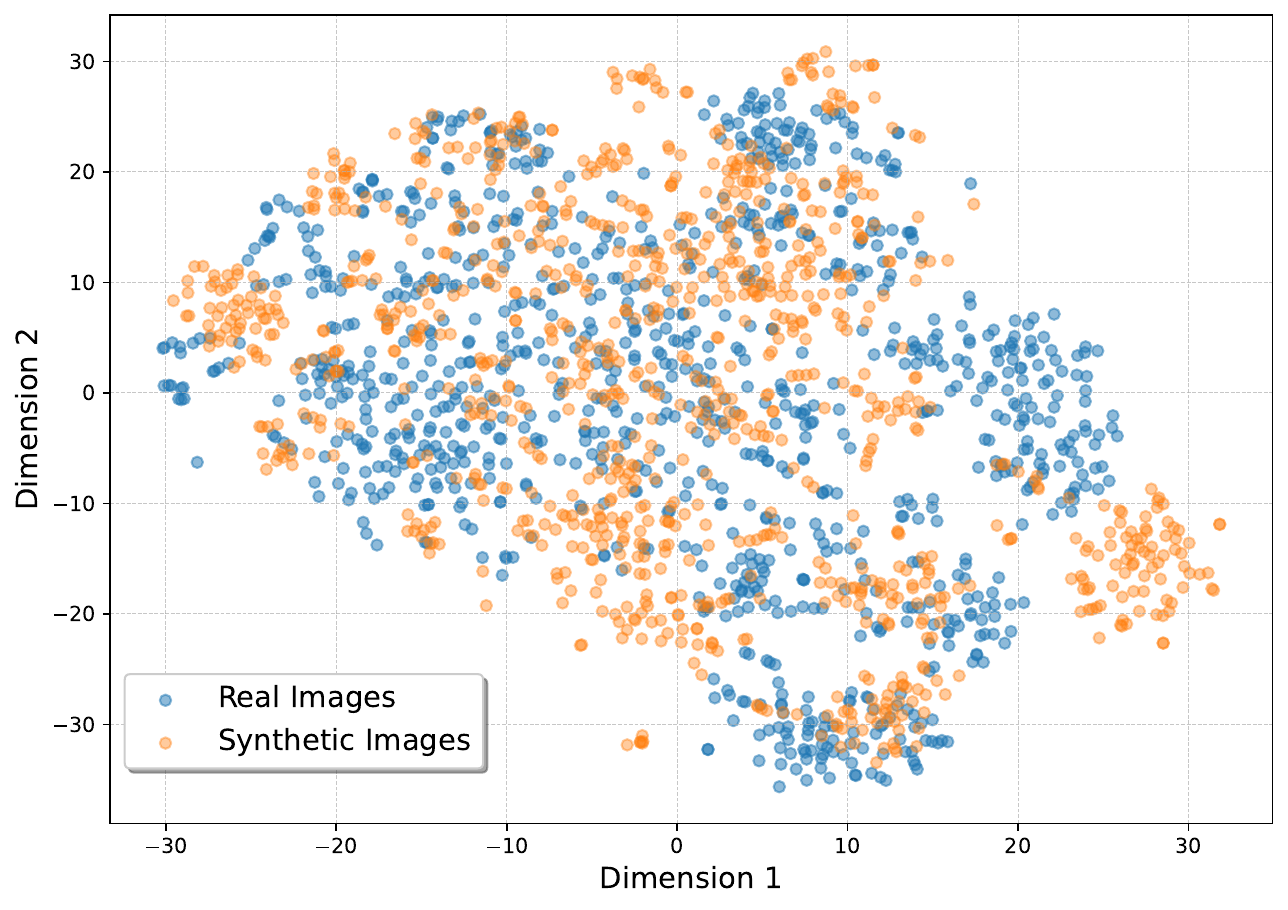}
    \caption{Visualization of synthetic and natural image features.} 
    
    \label{fig:Features_Visualization}
   
\end{figure}

\subsection{Effectiveness of Synthetic Images}

To validate the effectiveness of synthetic images in hallucination evaluation, we compared evaluation outcomes from three image sources under the same semantic distribution: a subset of COCO2014 images used in the POPE evaluation method, recent high-quality Internet images, and images generated via the ODE method with Stable-Diffusion 1.5. The results, presented in Fig.~\ref{fig:fig1}, indicate that test performance on COCO2014 images exceeds that on internet and ODE-generated images, suggesting potential data contamination, as the model may have encountered these images during prior training or fine-tuning. The observed difference in hallucination effects between ODE-generated images and internet images was minimal. 

Additionally, we produced synthetic images containing the same visual information as 1,004 natural images. We extracted features using the CLIP model and reduced them to a two-dimensional space, revealing a high degree of similarity between synthetic and natural images in the feature space, as shown in Fig.~\ref{fig:Features_Visualization}. These findings suggest that synthetic images, within an acceptable margin of error, are a viable and sustainable alternative for constructing open-set datasets. Furthermore, we anticipate that improvements in text-to-image models will enhance the quality of generated images.

\subsection{Main Results}
Table~\ref{table:evaluation} shows selected evaluation results (complete data are provided in the Appendix), and we found that:
\begin{itemize}
\item \textbf{Inconsistencies Between Static Benchmarks and ODE Performance:} Models like MiniGPT-4 and LLaVA perform well on static benchmarks, with F1-scores around 80-90. However, when tested with standard concept pairs in our evaluation, their scores declined. This suggests that these models may not fully understand fundamental concept relationships but rather rely on memorized correlations or noise from training data. This outcome highlights the limitations of static benchmarks, which can overestimate model capabilities due to data leakage and overfitting. Dynamic testing with diverse samples is necessary for a deeper evaluation of model learning and adaptability.

\begin{figure}[htb]
  \centering
  \centerline{\includegraphics[width=0.9\linewidth]{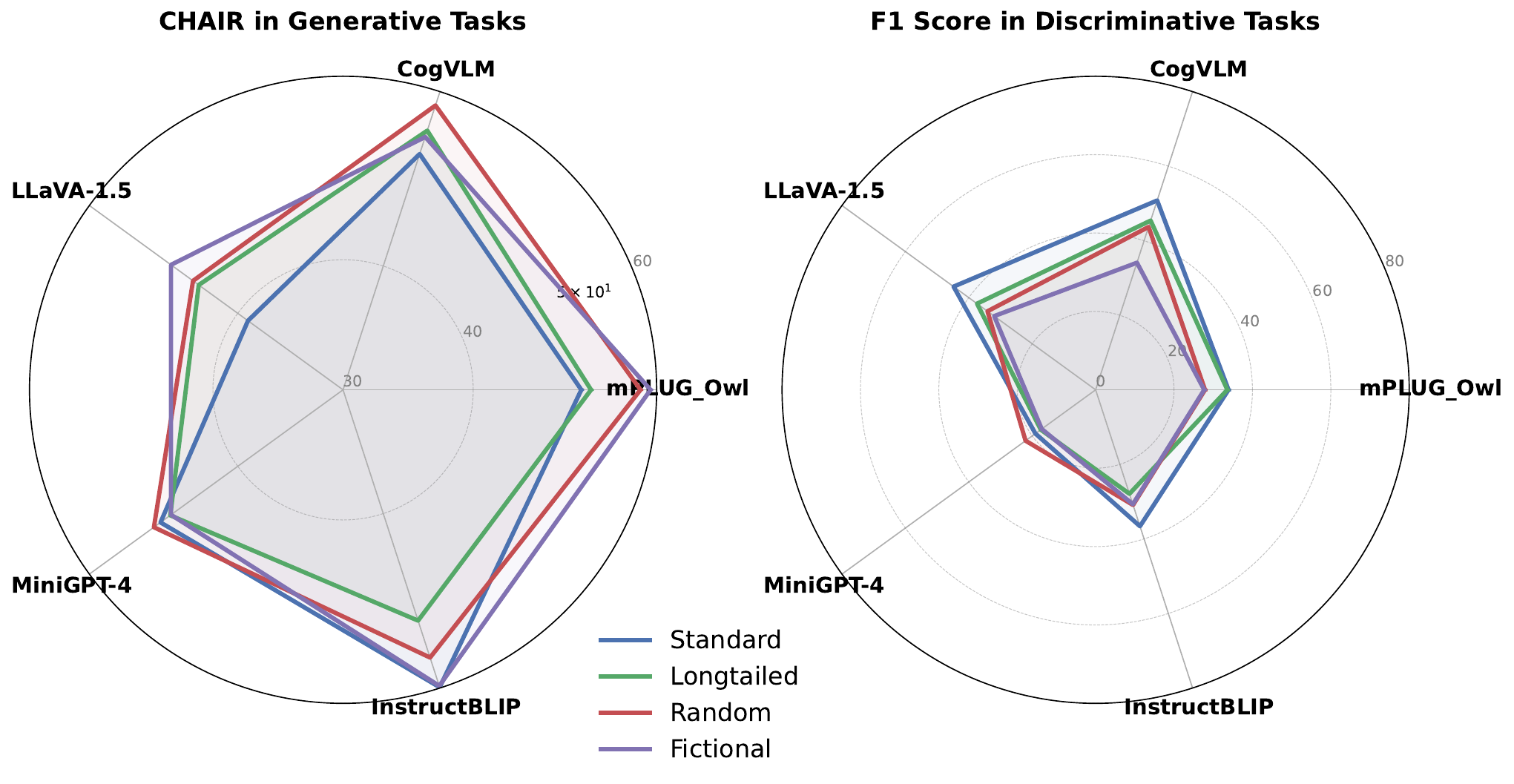}}
\caption{Performance across different distributions in generative tasks and discriminative tasks (existence-level).}
\label{fig:type-compare}
\end{figure}

\item \textbf{Distribution Range and Hallucinations:} Performance varies significantly across different distributions, as shown in Fig.~\ref{fig:type-compare}. On high-frequency concepts (e.g., Standard category), models perform well due to rich semantic associations. However, in object existence discrimination tasks, models like MiniGPT-4 exhibit instability, likely due to an over-reliance on high-frequency patterns, which hampers generalization. In the Random and Fictional categories,  hallucinations increase, particularly in attribute recognition, as the randomness of unrelated distributions amplifies generalization challenges and highlights the models' dependency on specific patterns. For low-frequency (Long-tailed) concepts, models perform adequately in generative tasks but struggle with object discrimination. This likely stems from insufficient low-frequency samples, limiting the foundational basis needed for generalized judgments and impairing the models' ability to handle combinations of low-frequency concept pairs.

\item \textbf{Different Requirements of Tasks on Semantic Understanding:} Our results reveal distinct adaptive differences in model performance between high-frequency and low-frequency concepts in generative vs. discriminative tasks. During the training phase, models are repeatedly exposed to these high-frequency combinations, establishing robust semantic associations. This is reinforced in generative tasks which emphasize semantic consistency. This flexibility enables generative models to produce outputs that broadly align with concepts, even with incomplete understanding of details. Discriminative tasks, however, require precise identification of target concepts, making models sensitive to noise or biases from overrepresented pairs in training data. This over-memorization affects accuracy more in discriminative tasks, where even small misclassifications disrupt coherence.

\item \textbf{Widening Performance Differences Among Models:} Different architectures and training methods, such as memory mechanisms and attention distributions, lead to performance variations. For example, mPLUG performs well in attribute discrimination, while MiniGPT-4 maintains balanced performance across tasks but lacks fine-grained attribute precision. Overall, LLaVA-1.5 exhibits robust performance across tasks and difficulty levels, while models like MiniGPT-4 and CogVLM highlight distinct strengths and weaknesses in data dependency and generalization capabilities.
\end{itemize}

\begin{table}[h]
\centering
\caption{Performance of models on both generative and discriminative tasks (object existence hallucination) across two datasets, with the difference ($\Delta$) between datasets.}
\setlength{\tabcolsep}{4pt} 
\renewcommand{\arraystretch}{1.0} 
\begin{adjustbox}{width=\linewidth,center}
\fontsize{9pt}{10pt}\selectfont
\begin{tabular}{ccccccc}
\toprule
\textbf{Model} & \textbf{Dataset} & \textbf{Accuracy} & \textbf{Precision} & \textbf{Recall} & \textbf{F1Score} \\
\midrule
\multirow{3}{*}{CogVLM} & ODE-Flux & 41.4  & 90.0  & 15.6  & 26.6 \\
                        & ODE-SD & 92.8  & 99.2  & 82.8  & 90.2 \\
                        & $\Delta$ & \cellcolor{lightgray} +51.4 & \cellcolor{lightgray} +9.2 & \cellcolor{lightgray} +67.2 & \cellcolor{lightgray} +63.6 \\
\midrule
\multirow{3}{*}{LLaVA-1.5} & ODE-Flux & 51.3 & 84.7 & 32.9 & 47.4 \\
                           & ODE-SD & 94.3  & 98.6  & 87.3  & 92.6 \\
                           & $\Delta$ & \cellcolor{lightgray} +43.0 & \cellcolor{lightgray} +13.9 & \cellcolor{lightgray} +54.4 & \cellcolor{lightgray} +45.2 \\
\midrule
\multirow{3}{*}{mPLUG\_Owl} & ODE-Flux & 38.1 & 91.4 & 7.9 & 14.5 \\
                       & ODE-SD & 66.1  & 86.5  & 20.4  & 33.0 \\
                       & $\Delta$ & \cellcolor{lightgray} +28.0 & \cellcolor{lightgray} -4.9 & \cellcolor{lightgray} +12.5 & \cellcolor{lightgray} +18.5 \\
\midrule
\multirow{3}{*}{MiniGPT-4} & ODE-Flux & 67.1 & 87.4 & 58.1 & 69.8 \\
                           & ODE-SD & 66.7  & 55.6  & 88.5  & 68.2 \\
                           & $\Delta$ & \cellcolor{lightgray} -0.4 & \cellcolor{lightgray} -31.8 & \cellcolor{lightgray} +30.4 & \cellcolor{lightgray} -1.6 \\
\midrule
\multirow{3}{*}{InstructBLIP} & ODE-Flux & 51.5 & 91.1 & 30.4 & 45.6 \\
                              & ODE-SD & 72.1  & 97.8  & 56.1  & 71.3 \\
                              & $\Delta$ & \cellcolor{lightgray} +20.6 & \cellcolor{lightgray} +6.7 & \cellcolor{lightgray} +25.7 & \cellcolor{lightgray} +25.7 \\
\bottomrule
\end{tabular}
\end{adjustbox}
\label{table:2performance}
\end{table}

\subsection{Dynamic Benchmarking Updates}
As a third-party benchmark evaluation, to ensure that models maintain robust performance in evolving application scenarios, we propose an ODE-based dynamic update mechanism for benchmark evaluations. This update mechanism involves the introduction of new pairs of target concepts, transformations in attribute combinations, and updates to the underlying knowledge base to include a broader range of object categories and their relationships. Such an iterative update process enables the benchmark to cover a wider range of distributional scenarios, ensuring both the accuracy and the comprehensiveness of model evaluation results. We conducted a second evaluation using data generated by Stable-Diffusion 1.5. Table~\ref{table:2performance} presents the model performance variations under a long-tail distribution. The significant discrepancies in performance indicate differing levels of model comprehension across various concepts.

\section{Methodological Applications}

\subsection{Hallucination Tendencies in Evaluation Results}

Our findings enable the analysis of hallucination tendencies within individual concepts and hallucination associations between different concepts. For example, we generated a frequency matrix of fact-hallucination concept pairs from LLaVA and performed clustering, identifying four groups, including indoor and traffic scene concepts. We found that hallucinations are more likely in scenarios with high contextual similarity or visual ambiguity.  For instance, clusters containing both indoor and outdoor concepts (e.g., ``car'' and ``chair'') exhibit higher hallucination rates, as shown in Table~\ref{table:3cluster_analysis}, revealing potential weaknesses in the model's understanding of scene context and object differentiation.

\begin{table}[h!]
\centering
\caption{Cluster Analysis: Top Truth Concepts}
\fontsize{9pt}{10pt}\selectfont
\begin{tabular}{p{3cm} p{4cm}}  
\toprule
\textbf{Cluster} & \textbf{Top Truth Concepts} \\
\midrule
Indoor Concepts & table, chair, floor, person, cat \\
Mixed Concepts & car, person, bird, chair, cluster \\
Traffic \& Outdoor & car, bench, bicycle, beach, road \\
Household Concepts & table, chair, cat, drink, lamp \\
\bottomrule
\end{tabular}
\label{table:3cluster_analysis}
\end{table}

\subsection{ODE in Domain-Specific Scenarios}
\begin{figure}[htb]
  \centering
  \centerline{\includegraphics[width=\linewidth]{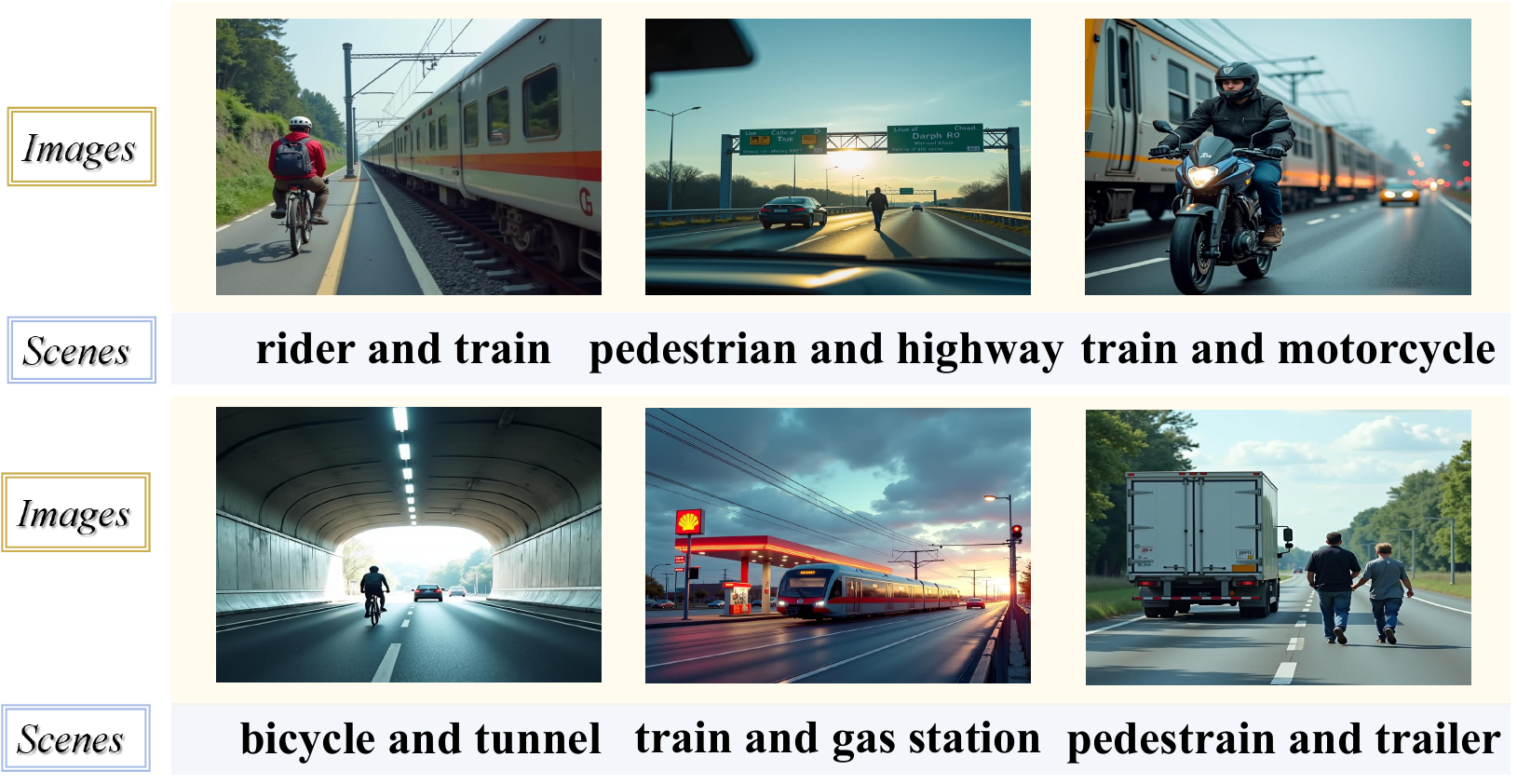}}
\caption{Examples of rare distribution samples constructed by our method in the transportation domain.}
\label{fig:traffic}
\end{figure}

Our framework also addresses hallucination detection in specific scenarios and data-scarce fields. The input flexibility of ODE allows for the selection of particular concepts, combined with its ability to generate customizable images, opening new possibilities for tailored data. As multimodal models expand into fields such as autonomous driving and healthcare, constructing diverse, challenging samples becomes critical to overcome existing datasets' limitations, such as narrow distributions and small sample sizes, which limit adaptability to complex scenarios. Fig.~\ref{fig:traffic} demonstrates our framework’s capability to generate rare concept combinations for the transportation domain, supplementing real-world scene data. This generative capability exposes models to scarce data during training, thereby enhancing real-world adaptability and mitigating overfitting risks. Additionally, fine-tuning with ODE-generated, distributionally diverse images improves model reliability in specialized domains.

\subsection{Fine-Tuning Enhancements}
Furthermore, fine-tuning MLLMs with ODE-generated data effectively mitigates object hallucinations. The ODE-generated test set avoids data contamination while offering rich distributional diversity, uncovering "thinking patterns" that drive hallucinations in specific scenarios. Based on this, targeted fine-tuning using the erroneous sample set can effectively mitigate the model's shortcomings. We utilize a dynamic, iteratively generated test set, which helps prevent model overfitting and the potential for "score boosting" on a single dataset. In our experiments, we extracted the erroneous samples exhibiting hallucinations from the initial round of generated data in the LLaVA model tests and performed fine-tuning on LLaVA. Subsequently, we evaluated the fine-tuned model using the existing AMBER test set. The results presented in Tables \ref{table:4generative} and \ref{table:discriminative_performance} indicate that the fine-tuned model shows significant improvements in performance across both generative and discriminative tasks.
\definecolor{deepblue}{RGB}{0, 102, 204}
\definecolor{brown}{RGB}{165, 42, 42}
\definecolor{lightgray}{gray}{0.9}
\definecolor{graytext}{gray}{0.4}

\begin{table}[htb]
\centering
\caption{Generative Task Performance Comparison}
\setlength{\tabcolsep}{4pt} 
\renewcommand{\arraystretch}{1.2} 
\begin{adjustbox}{max width=\linewidth,center}
\fontsize{9pt}{10pt}\selectfont
\begin{tabular}{>{\centering\arraybackslash}p{2.0cm}ccc>{\centering\arraybackslash}p{2cm}}
\toprule
\textbf{Model} & \textbf{CHAIR (\textcolor{brown}{$\downarrow$})} & \textbf{Cover (\textcolor{deepblue}{$\uparrow$})} & \textbf{Hal (\textcolor{brown}{$\downarrow$})} & \textbf{Cog (\textcolor{brown}{$\downarrow$})} \\
\midrule
Initial & 8.1 & 51.4 & 36.1 & 4.1 \\
Fine-tuned & \textbf{6.5} & 50.4 & \textbf{28.5} & \textbf{2.9} \\
\rowcolor{white} $\Delta$ & \cellcolor{lightgray} -1.6 & \cellcolor{lightgray} -1.0 & \cellcolor{lightgray} -7.6 & \cellcolor{lightgray} -1.2 \\
\bottomrule
\end{tabular}
\end{adjustbox}
\label{table:4generative}
\end{table}

\begin{table}[htb]
\centering
\caption{Discriminative Task Performance Comparison}
\label{table:discriminative_performance}
\setlength{\tabcolsep}{4pt} 
\renewcommand{\arraystretch}{1.2} 
\begin{adjustbox}{max width=\linewidth,center}
\fontsize{9pt}{10pt}\selectfont
\begin{tabular}{>{\centering\arraybackslash}p{2.0cm}cc>{\centering\arraybackslash}p{2cm}c}
\toprule
 & \multicolumn{2}{c}{Existence-level} & \multicolumn{2}{c}{Attribute-level} \\
\cmidrule(lr){2-5} 
\textbf{Model} & \textbf{Recall(\textcolor{deepblue}{$\uparrow$})} & \textbf{F1-Score(\textcolor{deepblue}{$\uparrow$})} & \textbf{Recall(\textcolor{deepblue}{$\uparrow$})} & \textbf{F1-Score(\textcolor{deepblue}{$\uparrow$})} \\
\midrule
Initial & 71.0 & 83.0 & 50.7 & 64.8 \\
Fine-tuned & \textbf{96.0} & \textbf{97.9} & \textbf{90.7} & \textbf{68.0} \\
\rowcolor{white}  $\Delta$ & \cellcolor{lightgray} +25.0 & \cellcolor{lightgray} +14.9 & \cellcolor{lightgray} +40.0 & \cellcolor{lightgray} +3.2 \\
\bottomrule
\end{tabular}
\end{adjustbox}

\end{table}

\begin{figure}[htb]
  \centering
  \centerline{\includegraphics[width=0.9\linewidth, height=4cm]{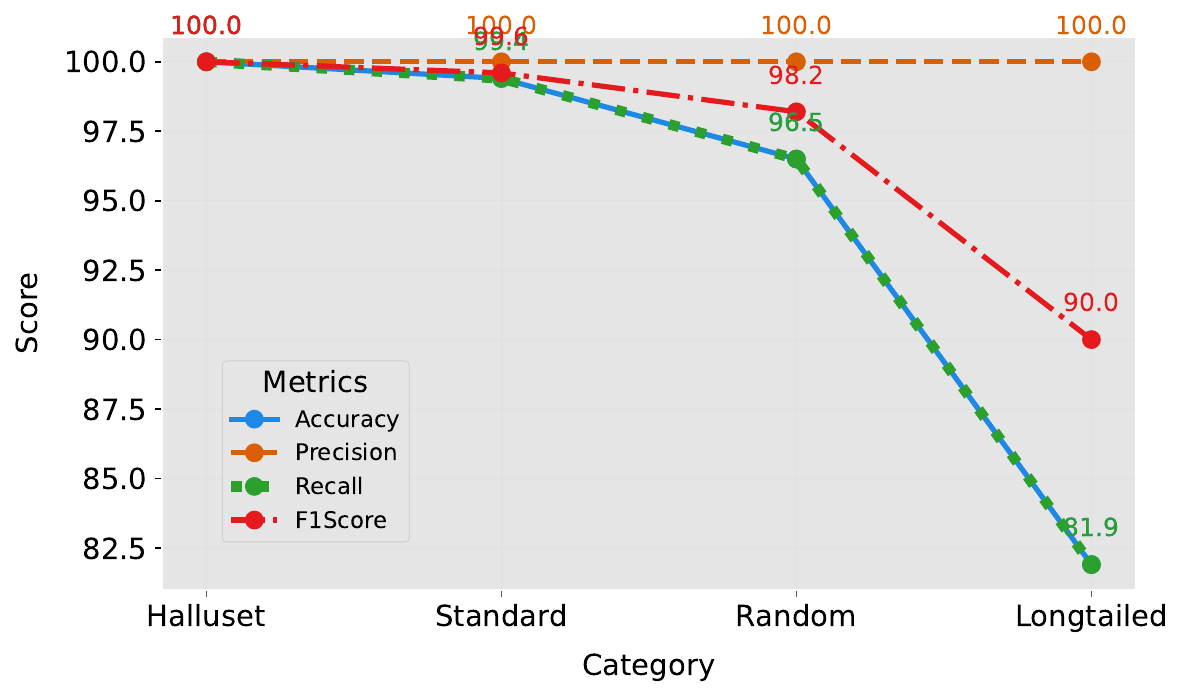}}
\caption{Performance of fine-tuned LLaVA models across distribution types, showing highest improvement with standard samples and limited impact from long-tail samples.}
\label{fig:metrics}
\end{figure}

The dataset, comprising Standard, Long-tail, Random, and Fictional distributions, enabled an open-set analysis. For the object existence hallucination task, 800 sample pairs from each distribution type were used for fine-tuning. The fine-tuned model exhibited better hallucination control. Fig.~\ref{fig:metrics} Results indicate that high-frequency distribution samples had the greatest positive effect, while long-tail samples contributed less than random samples. This discrepancy may arise from the regularity of high-frequency data, which facilitates model pattern recognition, whereas the imbalance in long-tail samples can lead to overfitting of frequent categories. Random samples provided better regularization, improving performance across diverse test sets. Incorporating diverse distributions in training data enhances robustness, generalization, and representation of low-frequency objects, boosting overall performance.

\section{Conclusions}

This paper addresses the issue of data contamination in the hallucination evaluation of multimodal large language models. We introduce a dynamic open-set evaluation protocol, initially applied to object hallucination at existence-level and attribute-level in visual question answering. The experimental results are more reliable than static benchmarks.
{
    \small
    \bibliographystyle{ieeenat_fullname}
    \bibliography{main}
}

\clearpage
\setcounter{page}{1}
\appendix

    
   

\section{Contamination and Synthetic Validity}
\begin{figure}[htb]
    \centering
    \includegraphics[width=\columnwidth, height=3.16cm]{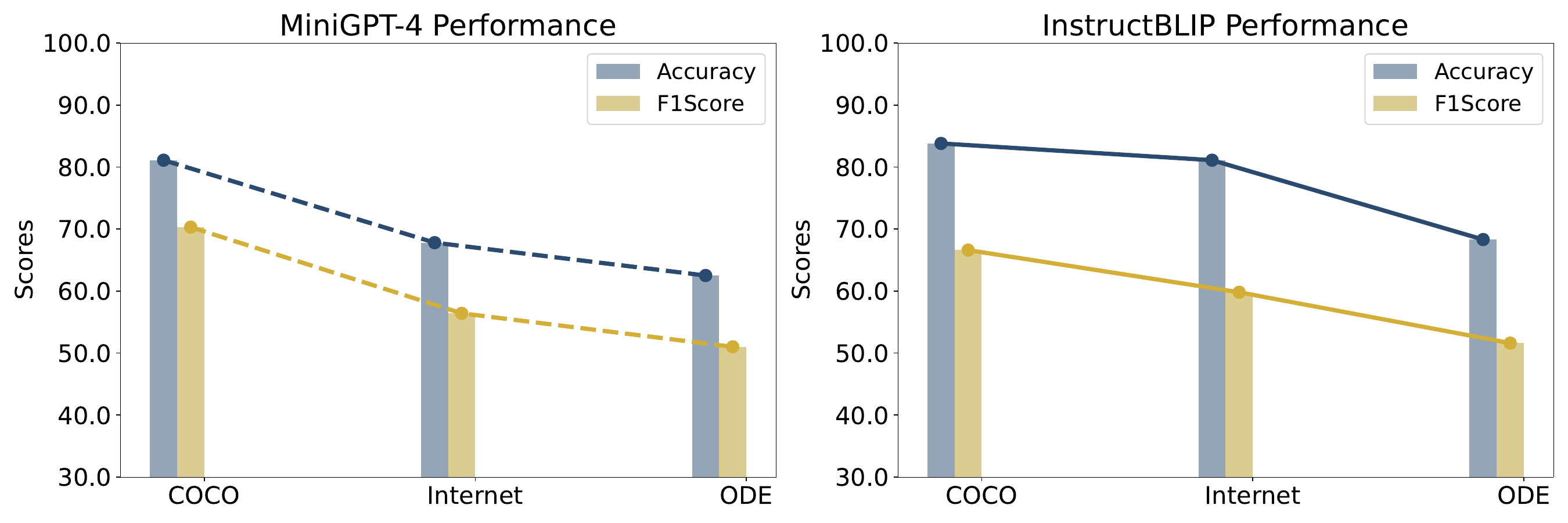}
    \caption{Model performances on discriminative tasks.} 
    
    \label{fig:ODE_performance}
   
\end{figure}

Fig.~\ref{fig:ODE_performance} highlights the performance of MiniGPT-4 and InstructBLIP on discriminative tasks across COCO 2014, Internet, and ODE-generated datasets. Internet data, as the most recent and reliably crawled dataset, avoids contamination, making it a more trustworthy baseline compared to COCO 2014, which shows artificially inflated results due to potential overlaps with training data. While ODE-generated synthetic data exhibits slightly lower performance than Internet data, feature space validates the close similarity between synthetic and natural images. This confirms that synthetic data can effectively replicate real-world distributions. Moreover, its controllability and distribution diversity establish it as a valuable resource for evaluating model reliability, particularly in mitigating data contamination and enabling the creation of novel and challenging testing scenarios.

The degradation results primarily indicate potential data contamination in certain models. Since the Internet and synthetic images do not share entirely identical semantics, some degree of discrepancy is expected due to differences in specific concepts within the images. 

\begin{wrapfigure}{r}{0pt}
    \vspace{-50pt}  
    \hspace{-25pt}  
    \includegraphics[width=0.1800\textheight]{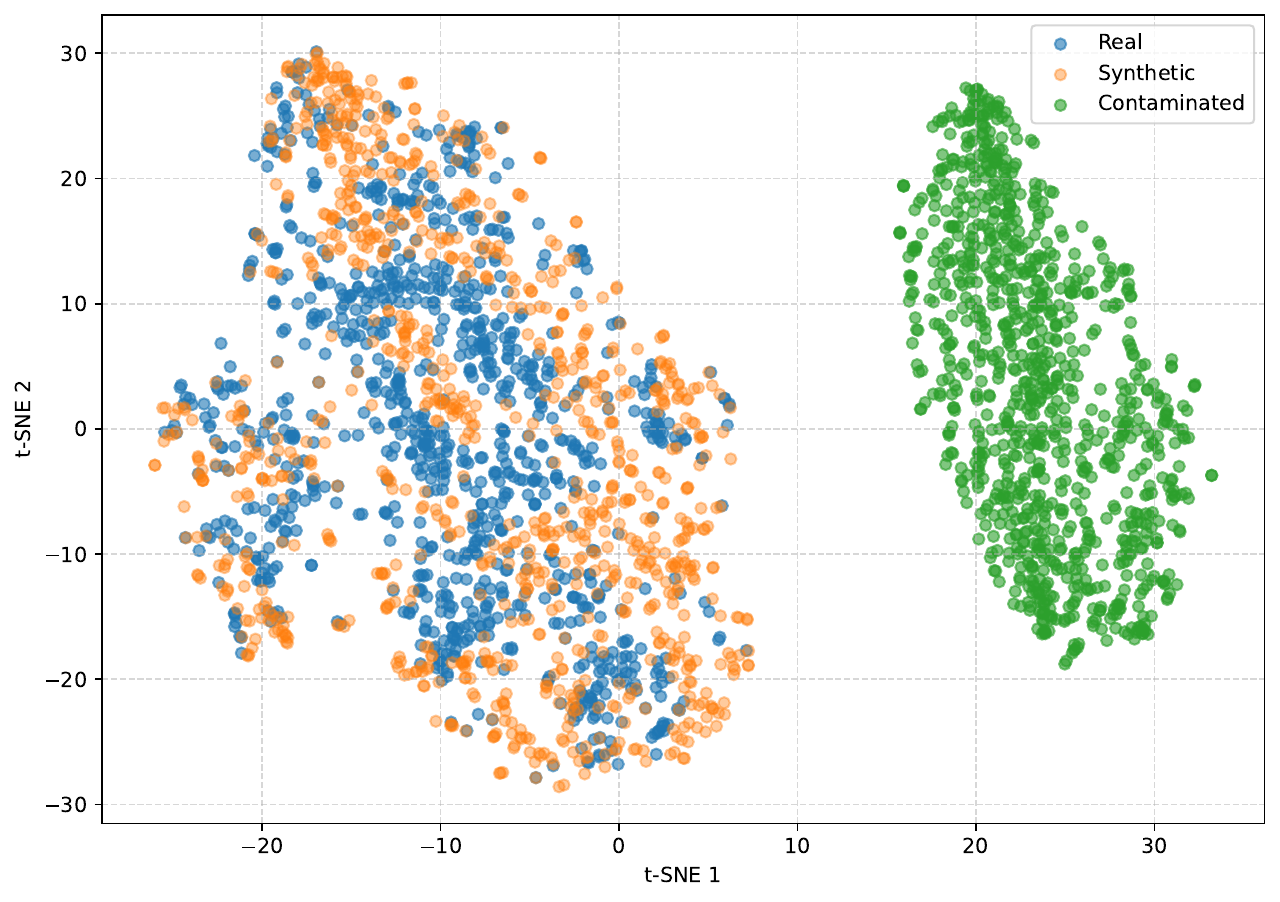}
    \captionsetup{justification=raggedright, font=small,}
    \caption{Feature Visualization.}
    \vspace{-15pt}
\end{wrapfigure}
We compared the features of three types of images: real images before contamination, contaminated images, and synthetic images generated on the basis of detailed semantic descriptions. The visualization shows that contaminated images form a distinct distribution, while synthetic images align closely with uncontaminated real images. Thus, we conclude that synthetic images are suitable for hallucination evaluation, a perspective also supported by previous works~\cite{HaloQuest}.

    
   

\section{Evaluation Results}

Table~\ref{tab:main_results_detail} summarizes the detailed evaluation results of five models under the following conditions: Standard, Random, Fictional, and Long-tail distributions. Tables~\ref{tab:table6}, \ref{tab:table7}, \ref{tab:table9}, and \ref{tab:table10} present the evaluation results of object-level hallucinations on ODE at the attribute level.

From the detailed results, we derive the following key observations:
\begin{itemize}
\item High Variability Across Attributes: Models excel in state-related attributes due to strong semantic associations in training data. However, action and number attributes expose significant weaknesses, especially in rare or unseen scenarios.

\item Impact of Data Distribution: Standard distributions enable high performance as a result of frequent exposure during training. In contrast, Fictional and Random distributions cause sharp performance declines, exposing over-reliance on memorized correlations. Long-tail distributions further highlight the models' inability to generalize effectively to sparse data.

\item Task-Specific Limitations:
Generative tasks emphasize semantic fluency, whereas discriminative tasks require detailed attribute recognition. These differences result in varying performance gaps between task types.

\item Model-Specific Trends: LLaVA-1.5 demonstrates the best overall performance with balanced precision and recall across attributes and distributions. InstructBLIP achieves high precision but suffers from poor recall, indicating a tendency toward overfitting. MiniGPT-4 and mPLUG-Owl struggle significantly with generalization, particularly in Random and Fictional contexts.

\item Broader Implications for Hallucination Mitigation:
Addressing relational hallucination requires improving the diversity and balance of training datasets. Attribute-specific fine-tuning and robust data augmentation strategies are essential for better generalization.
\end{itemize}

\definecolor{lightblue}{RGB}{200, 215, 235} 
\definecolor{deepblue}{RGB}{70, 110, 140} 
\definecolor{lightyellow}{RGB}{250, 235, 199} 
\definecolor{deepyellow}{RGB}{200, 150, 100} 
\definecolor{lightgray}{gray}{0.9}

\begin{table*}[ht]
    \centering
    \setlength{\tabcolsep}{9pt} 
    \caption{Evaluation results of five models (CogVLM, LLaVA-1.5, mPLUG\_Owl, MiniGPT-4, and InstructBLIP) under the following conditions: Standard, Random, Fictional, and Long-tail distributions. Each task type—Generative Task (CHAIR, Cover, Hal, Cog) and Discriminative Task (Existence and Attribute)—is evaluated with Accuracy (Acc), Precision (P), Recall (R), and F1-score (F1).}
    \renewcommand{\arraystretch}{1.2} 
    \begin{adjustbox}{max width=\textwidth,center}
    \fontsize{9pt}{10pt}\selectfont
    \begin{tabular}{l|l|p{0.7cm}p{0.7cm}p{0.7cm}p{0.7cm}|p{0.7cm}p{0.7cm}p{0.7cm}p{0.7cm}|p{0.7cm}p{0.7cm}p{0.7cm}p{0.7cm}}
        \toprule
        \multicolumn{2}{c|}{} & \multicolumn{4}{c|}{\textbf{Generative Task}} & \multicolumn{4}{c}{\textbf{Discriminative-Existence Task}} & \multicolumn{4}{|c}{\textbf{Discriminative-Attribute Task}} \\
         \cmidrule(r){3-6} \cmidrule(l){7-10} \cmidrule(l){11-14}
        \textbf{Criterion} & \textbf{Model} & \textbf{CHAIR} & \textbf{Cover} & \textbf{Hal} & \textbf{Cog} & \textbf{Acc} & \textbf{P} & \textbf{R} & \textbf{F1} & \textbf{Acc} & \textbf{P} & \textbf{R} & \textbf{F1} \\
        \midrule
        \multirow{5}{*}{Standard} 
        & CogVLM & 51.9 & 76.5 & 89.1 & 12.2 & 50.7 & \cellcolor{lightblue!30}\textbf{\textcolor{deepblue}{100.0}} & 26.2 & 41.5 & 55.2 & 46.8 & 55.6 & 50.8\\
        & LLaVA-1.5 & 38.9& \cellcolor{lightblue!30}\textbf{\textcolor{deepblue}{77.7}} & 82.7 & 8.6 & 69.5 & 97.8 & 55.4 & 70.7& 62.9 & 32.6 & 71.9 & 44.8\\
        & mPLUG & 50.8 & 77.2 & 96.0 & 11.5 & 41.7 & 94.7 & 13.4 & 23.5 & \cellcolor{lightblue!30}\textbf{\textcolor{deepblue}{72.5}} & 41.7 & 28.6 & 33.9\\
        & MiniGPT-4 & 49.4 & 76.0 & 93.6 & 14.2 & 64.5 & 97.5 & 48.0 & 64.3 & 69.5 & 39.7 & 12.5 & 19.0\\
        & InstructBLIP & 59.9 & 75.7 & 88.1 & 11.0 & 66.7 & 96.8 & 51.7 & 67.4 & 60.8 & 28.5 & 51.2 & 36.6\\
         & InternVL-2.5 & 38.5	&60.9	&\cellcolor{lightyellow!30}\textbf{\textcolor{deepyellow}{72.3}}&	\cellcolor{lightyellow!30}\textbf{\textcolor{deepyellow}{4.0}}	&75.7	&96.1	&66.3	&78.4	&65.9&	48.1&	87.0	&61.9\\
        & Qwen2-VL & 44.2	&75.2	&88.6	&11.6&	75.2&	97.0	&64.9	&77.7	&67.0	&\cellcolor{lightblue!30}\textbf{\textcolor{deepblue}{49.6}}	&86.8&	\cellcolor{lightblue!30}\textbf{\textcolor{deepblue}{63.1}}\\  
        & Cambrian &\cellcolor{lightyellow!30}\textbf{\textcolor{deepyellow}{36.3}} 	&64.9	&79.2	&7.3&	\cellcolor{lightblue!30}\textbf{\textcolor{deepblue}{76.9}}	&96.5&	\cellcolor{lightblue!30}\textbf{\textcolor{deepblue}{67.8}}&	\cellcolor{lightblue!30}\textbf{\textcolor{deepblue}{79.6}}&	65.7	&47.7	&\cellcolor{lightblue!30}\textbf{\textcolor{deepblue}{90.4}}&	62.4\\ 
        \midrule
        \multirow{5}{*}{Random} 
        & CogVLM & 58.1 & 57.7 & 87.6 & 6.0 & 40.0 & 89.7 & 18.0 & 30.0 & 57.1 & 82.5 & 38.2 & 52.2 \\
        & LLaVA-1.5 & 45.2 & 57.7 & 84.2 & 4.7 & 74.7 & 89.7 & 69.6 & 78.3 & \cellcolor{lightblue!30}\textbf{\textcolor{deepblue}{75.7}} & 90.1 & 65.9 & \cellcolor{lightblue!30}\textbf{\textcolor{deepblue}{76.1}}\\
        & mPLUG & 57.9 & 56.4 & 92.1 & 6.3 & 40.0 & 84.0 & 10.8 & 19.1 & 48.4 & 76.5 & 20.2 & 31.9\\
        & MiniGPT-4 & 50.3 & 57.9& 82.2& 5.8 & 66.4 & 86.9 & 58.2 & 69.7 & 45.0 & 74.5 & 10.8 & 18.8 \\
        & InstructBLIP & 55.9 & 58.4 & 83.2 & 5.6 & 64.6 & 87.6 & 54.6 & 67.2 & 72.3 & \cellcolor{lightblue!30}\textbf{\textcolor{deepblue}{91.5}} & 58.6 & 71.4\\
        & InternVL-2.5 & 43.0&	58.4&	67.8&	3.7&	81.4&	\cellcolor{lightblue!30}\textbf{\textcolor{deepblue}{89.9}}	&80.7	&85.0&	72.8&	67.0&	80.6&	73.1\\
        & Qwen2-VL & 38.5&	\cellcolor{lightblue!30}\textbf{\textcolor{deepblue}{60.9}} 	&72.3&	4.0	&\cellcolor{lightblue!30}\textbf{\textcolor{deepblue}{82.8}}&	88.8&	\cellcolor{lightblue!30}\textbf{\textcolor{deepblue}{83.5}}	&\cellcolor{lightblue!30}\textbf{\textcolor{deepblue}{86.0}}&	71.4	&65.8&	\cellcolor{lightblue!30}\textbf{\textcolor{deepblue}{82.6}}&73.2\\  
        & Cambrian &\cellcolor{lightyellow!30}\textbf{\textcolor{deepyellow}{30.9}} &	56.9	&\cellcolor{lightyellow!30}\textbf{\textcolor{deepyellow}{61.4}} &	\cellcolor{lightyellow!30}\textbf{\textcolor{deepyellow}{2.3}}&	80.7&	87.0&	83.0	&84.9	&71.2&	63.4	&84.3&72.3\\ 
        \midrule
         \multirow{5}{*}{Fictional} 
        & CogVLM & 54.0 & 55.9 & 80.7 & 6.3 & 39.5 & 90.9 & 16.1 & 27.3 & 50.5 & 75.4 & 28.7 & 41.5 \\
        & LLaVA-1.5 & 48.0 & 54.2 & 79.2 & 4.8 & 72.5 & 87.9 & 66.7 & 75.8 & \cellcolor{lightblue!30}\textbf{\textcolor{deepblue}{73.0}} & 88.5 & 62.0 & \cellcolor{lightblue!30}\textbf{\textcolor{deepblue}{72.9}}\\
        & mPLUG & 59.2 & 54.0 & 90.6 & 6.6 & 42.5 & \cellcolor{lightblue!30}\textbf{\textcolor{deepblue}{95.7}} & 11.8 & 21.0 & 45.8 & 72.1 & 17.4 & 28.0\\
        & MiniGPT-4 & 48.0 & 55.4 & 79.2 & 7.3 & 67.1 & 87.4 & 58.1 & 69.8 & 42.9 & 75.0 & 9.7 & 17.2\\
        & InstructBLIP & 59.7 & 56.9 & 80.2 & 5.9 & 66.0 & 88.6 & 56.2 & 68.7 & 71.0 & \cellcolor{lightblue!30}\textbf{\textcolor{deepblue}{88.8}} & 58.4 & 70.4\\
        & InternVL-2.5 &43.7 &	55.4 &	65.3 &	4.5 &	78.6 &	86.5 &	79.3 &	82.7 &	66.2 &	58.0 &	73.7 &	64.9\\
        & Qwen2-VL &41.5	 &\cellcolor{lightblue!30}\textbf{\textcolor{deepblue}{60.4}} 	 &69.8	 &5.3 &	77.2	 &86.6	 &76.6	 &81.2	 &65.8	 &59.0 &	75.0 &	66.0\\  
        & Cambrian &\cellcolor{lightyellow!30}\textbf{\textcolor{deepyellow}{34.5}}  &	55.0 &\cellcolor{lightyellow!30}\textbf{\textcolor{deepyellow}{63.4}}&	\cellcolor{lightyellow!30}\textbf{\textcolor{deepyellow}{3.4}} &\cellcolor{lightblue!30}\textbf{\textcolor{deepblue}{79.1}}	 &86.6 &	\cellcolor{lightblue!30}\textbf{\textcolor{deepblue}{80.1}} &\cellcolor{lightblue!30}\textbf{\textcolor{deepblue}{83.2}} &	67.1 &	58.3 &	\cellcolor{lightblue!30}\textbf{\textcolor{deepblue}{77.2}} &66.4\\ 
        \midrule
        \multirow{5}{*}{Long-tail} 
        & CogVLM & 54.8 & 67.1 & 89.6 & 14.3 & 41.4 & 90.0 & 15.6 & 26.6 & 61.5 & 77.3 & 49.0 & 59.9 \\
        & LLaVA-1.5 & \cellcolor{lightyellow!30}\textbf{\textcolor{deepyellow}{44.5}} & 71.3 & 91.1 & 11.2 & 51.3 & 84.7 & 32.9 & 47.4 & \cellcolor{lightblue!30}\textbf{\textcolor{deepblue}{78.2}} & \cellcolor{lightblue!30}\textbf{\textcolor{deepblue}{85.5}} & 74.3 & \cellcolor{lightblue!30}\textbf{\textcolor{deepblue}{79.5}}\\
        & mPLUG & 51.9 & 71.0 & 96.0 & 10.7& 38.1 & 91.4 & 7.9 & 14.5 & 48.9 & 67.4 & 24.8 & 36.2\\
        & MiniGPT-4 & 48.1 & \cellcolor{lightblue!30}\textbf{\textcolor{deepblue}{75.2}} & 93.1 & 16.5 & 63.4 & \cellcolor{lightblue!30}\textbf{\textcolor{deepblue}{92.9}} & 48.8 & 63.9& 48.0 & 75.9 & 17.1 & 27.9 \\
        & InstructBLIP & 51.3 & 71.0 & 87.1 & 11.2 & 51.5 & 91.1 & 30.4 & 45.6 & 73.3 & 84.6 & 65.1 & 73.5\\
        & InternVL-2.5 &50.4	&70.3	&86.1&	11.1	&\cellcolor{lightblue!30}\textbf{\textcolor{deepblue}{70.1}}&	88.1&	63.9	&74&	65.2	&50.5&	\cellcolor{lightblue!30}\textbf{\textcolor{deepblue}{89.7}}&	64.6\\
        & Qwen2-VL &49.8	&72.3&	93.1&	16.7&	69.3	&92.2	&58.9	&\cellcolor{lightblue!30}\textbf{\textcolor{deepblue}{71.8}}&	61.4&	48.2&	82.9&	60.9\\  
        & Cambrian &45.0	&53.0&	\cellcolor{lightyellow!30}\textbf{\textcolor{deepyellow}{80.7}}&	\cellcolor{lightyellow!30}\textbf{\textcolor{deepyellow}{9.0}}&	68.5&	89.6&	\cellcolor{lightblue!30}\textbf{\textcolor{deepblue}{59.7}}	&71.6&	60.8&	47.0&	88.7&	61.4\\ 
        \bottomrule
    \end{tabular}%
    \end{adjustbox}
  
    \label{tab:main_results_detail}
\end{table*}

\begin{table}[ht]
\centering
\caption{The detailed evaluation results of ODE on the object hallucination (attribute-level) under Standard distribution.}
\begin{adjustbox}{max width=\linewidth}
\small 
\begin{tabular}{p{0.1\linewidth}p{0.1\linewidth}cccccc} 
\toprule
 & Metric & mPLUG-Owl & MiniGPT-4 & LLaVA-1.5 & CogVLM & InstructBLIP \\
\midrule
\multirow{4}{*}{State} & Acc & 57.8 & 51.3 & \textbf{80.5} & 69.5 & 71.3 \\
 & P   & 75.3 & 84.2 & 86.8 & \textbf{87.0} & 86.8 \\
 & R   & 28.6 & 12.5 & \textbf{71.9} & 55.6 & 51.2 \\
 & F1  & 41.4 & 21.7 & \textbf{78.6} & 67.8 & 64.4 \\
\midrule
\multirow{4}{*}{Number} & Acc & 40.4 & 39.9 & \textbf{92.6} & 54.5 & 87.0 \\
 & P   & 82.7 & 93.5 & 80.2 & 98.4 & \textbf{98.5} \\
 & R   & 15.3  & 10.6 & \textbf{90.3} & 32.7 & 81.9 \\
 & F1  & 25.8 & 19.0 & \textbf{94.1} & 49.1 & 89.4 \\
\midrule
\multirow{4}{*}{Action} & Acc & 53.3 & 43.9 & \textbf{66.9} & 60.1 & 67.3 \\
 & P   & 67.8 & 45.8 & \textbf{74.2} & 66.9 & 74.5 \\
 & R   & 17.5 & 9.6 & \textbf{51.8} & 51.3 & 52.6 \\
 & F1  & 27.8 & 15.8 & 61.0 & 58.0 & \textbf{61.6} \\
\bottomrule
\end{tabular}
\end{adjustbox}
\label{tab:table6}
\end{table}

\begin{table}[htb]
\centering
\caption{The detailed evaluation results of ODE on the object hallucination (attribute-level) under Long-tail distribution.}
\begin{adjustbox}{max width=\linewidth}
\small 
\begin{tabular}{p{0.1\linewidth}p{0.1\linewidth}cccccc} 
\toprule
 & Metric & mPLUG-Owl & MiniGPT-4 & LLaVA-1.5 & CogVLM & InstructBLIP \\
\midrule
\multirow{4}{*}{State} & Acc &55.7&	50.5	&\textbf{75.5}&	66.1	&64.2
 \\
 & P   & 69.1	&71.4&	82.5&	\textbf{84.3}&	74.6
\\
 & R   & 32.2&	13.7	&\textbf{64.7}	&53.4&	43.2
\\
 & F1  & 43.9&	23.0&	\textbf{72.5}&	65.3	&54.7
\\
\midrule
\multirow{4}{*}{Number} & Acc & 40.1&	42.4	&\textbf{86.8}	&49.8&	83.7
 \\
 & P   & 78.9	&85.7	&95.5&	91&\textbf{	97.2}
\\
 & R   & 14.9	&16.3	&\textbf{94.2}	&27.5&	78
 \\
 & F1  & 25&	27.44&	\textbf{89.4}&	42.2	&86.5
\\
\midrule
\multirow{4}{*}{Action} & Acc & 48.5&	51.9	&\textbf{64.4}&	61	&\textbf{64.4}
 \\
 & P   & 50.7&	62.8	&64.2	&61.8	&\textbf{64.6}
 \\
 & R   & 27.3&	20.5&	65.2&	\textbf{71.2}&	63.6
 \\
 & F1  & 35.4&	30.9&	64.6&	\textbf{66.1}&	64.0
 \\
\bottomrule
\end{tabular}
\label{tab:table7}
\end{adjustbox}
\end{table}

\begin{table}[ht]
\centering
\caption{The detailed evaluation results of ODE on the object hallucination (attribute-level) under Fictional distribution.}
\begin{adjustbox}{max width=\linewidth}
\small 
\begin{tabular}{p{0.1\linewidth}p{0.1\linewidth}cccccc} 
\toprule
 & Metric & mPLUG-Owl & MiniGPT-4 & LLaVA-1.5 & CogVLM & InstructBLIP \\
\midrule
\multirow{4}{*}{State} & Acc & 51.3 & 50.2 & \textbf{71.1} & 55.0 & 66.4 \\
 & P   & 67.5 & 76.5 & \textbf{87.7} & 84.4 & 85.7 \\
 & R   & 22.4 & 11.2 & \textbf{49.1} & 28.0 & 41.4 \\
 & F1 & 33.6	&19.5&	\textbf{62.9}	&42&	55.8 \\
\midrule
\multirow{4}{*}{Number} & Acc & 39.9	&35.8	&\textbf{76.9}&	46&	73.4 \\
 & P   & 85.7&	72.7	&91.5&86.6	&\textbf{93.6} \\
 & R   & 13.4&	5.9	&\textbf{72.0}	&24.0	&65.1 \\
 & F1  & 23.2	&10.9&	\textbf{80.5}	&37.6&	76.7 \\
\midrule
\multirow{4}{*}{Action} & Acc & 45.7&	46.7&	\textbf{56.5}&	42.4&	57.6
 \\
 & P   &30.0&	\textbf{90.9}&	60. 0&	46.5&	62.5&
\\
 & R   & 6.5&	21.7&	39.1&	\textbf{43.5}&	\textbf{43.5}
 \\
 & F1  & 10.7&	35.0	&47.3	&44.9	&\textbf{51.2}
 \\
\bottomrule
\end{tabular}
 \vspace{-10mm}
\end{adjustbox}
\label{tab:table9}
\end{table}

\begin{table}[ht]
\centering
\caption{The detailed evaluation results of ODE on the object hallucination (attribute-level) under Random distribution.}
\begin{adjustbox}{max width=\linewidth}
\small 
\begin{tabular}{p{0.1\linewidth}p{0.1\linewidth}cccccc} 
\toprule
 & Metric & mPLUG-Owl & MiniGPT-4 & LLaVA-1.5 & CogVLM & InstructBLIP \\
\midrule
\multirow{4}{*}{State} & Acc & 55.0&	53.5    &	\textbf{71.9}&	60.3&	66.1
 \\
 & P   & 76.4	&78.3	&86.3&	85.0	&\textbf{87.5}
 \\
 & R   & 22.7	&14.9	&\textbf{52.1}&	39.7&	40.5
 \\
 & F1  & 35.0	&25.0	&\textbf{64.9}&	54.1&	55.3
 \\
\midrule
\multirow{4}{*}{Action} & Acc & 40.9&	36.8	&\textbf{80.4}&	48.0&	75.2
 \\
 & P   &79.8&	78.4&	92.5&	87.2&	\textbf{95.0}
 \\
 & R   & 16.6&	7.2	&\textbf{76.7}&	27.0	&66.8
 \\
 & F1  & 27.5&	13.2	&\textbf{83.8}	&41.2&	78.4
 \\
 \midrule
\multirow{4}{*}{Number} & Acc &54.7&	45.3&	54.7&	65.6	&\textbf{60.9}
 \\
 & P   & 75&	0	&75	&\textbf{76.9}&	\textbf{76.9}
 \\
 & R   & 18.7	&0&	18.7&	\textbf{62.5}&	31.2
 \\
 & F1  & 29.9	&0&	29.9&	\textbf{68.9}&	44.3
 \\

\bottomrule
\end{tabular}
 \vspace{-10mm}
\end{adjustbox}
\label{tab:table10}
\end{table}

\end{document}